%% file: root.tex
\pdfoutput=1
\documentclass[letterpaper, 10 pt, conference]{ieeeconf}  

\IEEEoverridecommandlockouts                              

\usepackage{graphicx}
\graphicspath{{../pdf/}{../jpeg/}}
\usepackage{subfigure}
\usepackage{hyperref}
\usepackage{microtype}
\usepackage{makecell}
\usepackage{amsmath}
\usepackage{bm}
\usepackage{color}
\usepackage{multirow}
\usepackage{algorithm}
\usepackage{algorithmic}
\usepackage{etoolbox}
\apptocmd{\thebibliography}{\raggedright}{}{}
\usepackage{flushend}

\title{Collision-free Trajectory Planning for Autonomous Surface Vehicle}

\author{Licheng~Wen $^{1}$ ,  
  Jiaqing~Yan $^{1}$   ,  
  Xuemeng~Yang $^{1}$  ,
  Yong~Liu $^{1,2}$  ,and
  Yong~Gu$^{1}$
\thanks{$^{1}$Licheng~Wen , Jiaqing~Yan , Xuemeng~Yang ,Yong~Liu and Yong~Gu are from College of Control
Science and Technology, Zhejiang University,Hangzhou, China }%
\thanks {$^{2}$Yong Liu is with the State Key Laboratory of Industrial Control Technology and Institute of Cyber-Systems and Control, Zhejiang University, Zhejiang, 310027, China (Yong Liu is the corresponding author, {\tt\small yongliu@iipc.zju.edu.cn})}
}
\begin{document}

\maketitle
\thispagestyle{empty}
\pagestyle{empty}


\begin{abstract}
    In this paper, we propose an efficient and accurate method for autonomous surface vehicles to generate a smooth and collision-free trajectory considering its dynamics constraints. We decouple the trajectory planning problem as a front-end feasible path searching and a back-end kinodynamic trajectory optimization. Firstly, we model the type of two-thrusts under-actuated surface vessel. Then we adopt a sampling-based path searching to find an asymptotic optimal path through the obstacle-surrounding environment and extract several waypoints from it. We apply a numerical optimization method in the back-end to generate the trajectory. From the perspective of security in the field voyage, we propose the sailing corridor method to guarantee the trajectory away from obstacles. Moreover, considering limited fuel ASV carrying, we design a numerical objective function which can optimize a fuel-saving trajectory. Finally, we validate and compare the proposed method in simulation environments and the results fit our expected trajectory. 
\end{abstract}


\input{section/01_Introduction.tex}
\input{section/02_Relatedwork.tex}
\input{section/03_Methodology.tex}
\input{section/04_TrajectoryGeneration.tex}
\input{section/05_ExperimentAndResult.tex}
\input{section/06_Conclusion.tex}


\section*{Acknowledgement}
This work is supported by the National Key R\&D program of China under Grant 2018YFB1305900 and the National Natural Science Foundation of China under Grant 61836015.

%
%


\bibliographystyle{spmpsci}      
\bibliography{cite.bib}

\end{document}

%% file: section/01_Introduction.tex
\section{Introduction}
%
%
%
%

Autonomous surface vehicles(ASVs) motivated by its cheapness and eco-friendly operation, has sparked off much concern in the realm of robotics. Known for its reliable safety and efficiency, the ASV can meet the demands for fast, unmanned, technically reasonable multi-missions\cite{yan2010development}, including hydrological surveys, environment monitoring, and maritime rescuing.

The trajectory planning problem is not novel to researchers. As the traditional graph-based and sampling-based path searching methods \cite{hao2005planning} \cite{lavalle2001randomized} \cite{karaman2011sampling} output position-only path, ASVs find it hard to follow due to lacking of directions and velocities on path. The generation of kinodynamic feasible trajectory requires taking ASV's nonlinear and under-actuated system into account. \cite{hervagault2017trajectory} first provides a differential flatness description of ASV's dynamics. \cite{louembet2010motion} applies B-spline parameterization to deal with the dynamical constraints. \cite{mellinger2011minimum} optimizes the trajectory by minimizing a cost function derived from the snaps and accelerations. 

Although plenty of works on ASV trajectory planning have been proposed, there are still two critical issues that few researchers have mentioned. Firstly, there are usually obstacles in the target area. However, previous researches fail to focus on that. Besides, those who take obstacles into accounts rarely perform their trajectories in simulation environments or field experiments either. Secondly, energy cost plays a key role in our issue since ASVs carrying limited fuel out for missions. Researchers have considered some factors as optimized objectives in trajectory planning, such as accelerations, jerks, and path smoothness, but never concentrate on fuel-saving.

In this paper, we propose a complete trajectory planning method to address these two issues systematically. Based on the modeling of ASV with two thrusts, we propose a framework that decouples the problem into front-end path searching and back-end trajectory optimization. We propose a method called ``sailing corridor'' to guarantee the trajectory both collision-free and kinodynamic feasible. 

As for the energy-saving factor, fuel consumption is affected directly by the pumps' thrusts, which is also our control input. By setting minimum control input as the optimization objective function, we can obtain a fuel-saving trajectory.

The contributions of the present work are:
\begin{itemize}
    \item Propose a method to generate smooth and collision-free trajectories using sailing corridor constraints; 
    \item Design a numerical objective function which can optimize a fuel-saving trajectory; 
    \item Perform experiments in simulation to verify our method's efficiency and accuracy;
\end{itemize}

\begin{figure}[!t]
\centering
\includegraphics[width=0.95\linewidth]{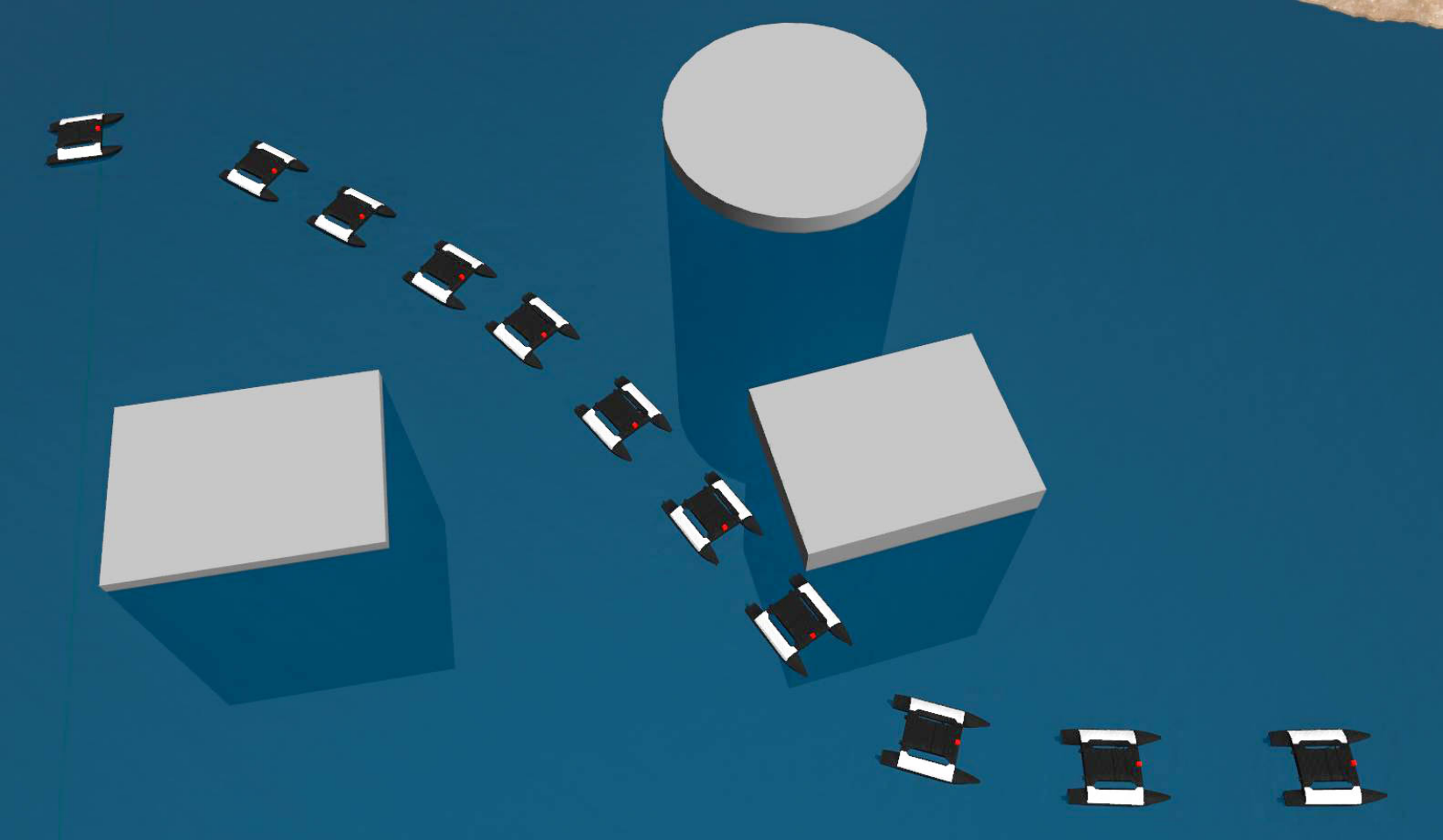}
\caption{Composite images of Kingfisher model sailing in the Gazebo simulation environment}
\label{simu}
\end{figure}

The paper is organized as follows. Section \uppercase\expandafter{\romannumeral2} reviews the previous work related to our problem.  Section \uppercase\expandafter{\romannumeral3} presents the ASV dynamic model and the front-end/back-end methodology. Section \uppercase\expandafter{\romannumeral4} presents trajectory optimization problem with different objective functions, including fuel-saving. We also introduce constraints to make the trajectory smooth and collision-free. In Section \uppercase\expandafter{\romannumeral5}, the proposed solution is evaluated and compared with published methods in a simulation environment. Finally, Section \uppercase\expandafter{\romannumeral6} summarizes and concludes the presented work.

%% file: section/02_Relatedwork.tex
\section{Related Work}

\textit{Path searching} is to find a path in the configuration space of the vehicle from the initial location to the target region. 
This problem can be solved by several available methods, which produces feasible or optimal waypoints, and those methods are divided into two main aspects: graph-search based methods and incremental sampling-based ones. 
The first one describes the map as an occupancy grid or lattice where vertices represent a collection of reachable configurations, and edges represent links between vertices.
The Dijkstra algorithm presented in \cite{lavalle1998optimal} is a typical graph searching method, which discretizes the known map to cell-grid space and adds weights for each cell to find the shortest path.
\cite{hao2005planning} first proposes a widely-used A* algorithm for unmanned vehicles, to produce an optimal path due to the implementation of heuristics, but usually costs much computational time and memory.
\cite{montemerlo2008junior} proposes the hybrid A* algorithm, which implemented the A* with practical constraints and performed well in DARPA Urban Challenge.
\cite{niu2018energy} proposes an energy-efficient method combining Dijkstra's search, energy consumption, and Voronoi-Visibility. It can perform well in terms of the path's smoothness and continuity but does not address dynamic constraints. 
\cite{pivtoraiko2005efficient} discretizes the planning area with grids with states, called state lattice, where the path searching methods applied. It produces a smooth path for vehicles and performs well in relatively simple environments.
The latter one randomly samples the configuration space or state space to find a feasible path, dealing with high dimensional spaces well with time constraints.
\cite{lavalle2001randomized} proposes a commonly used sampling-based algorithm, rapidly-exploring random tree(RRT), which builds a random tree of trajectories to find a path and is applied to dynamic systems. However, it generates jerky or not curvature continuous trajectories under some circumstances.
\cite{karaman2011sampling} proposes RRT*,  which iteratively builds a tree and lowers the given cost function through the state space.  It is proved to be both asymptotic optimal and computationally efficient but still has the same deficiency as RRT.
\cite{karaman2010optimal} extends RRT* to deal with differential-constraints models, applied well in anytime motion planning for Dubins' vehicles.

\textit{Trajectory generation} used after path and waypoints found, ensures the trajectory smooth and collision-free, taking the vehicle's dynamic constraints into account. 
The traditional approaches to this problem can also be divided into two aspects: the interpolating curve approaches and the numerical optimization methods \cite{gonzalez2015review}.  
The first one replaces non-smooth pieces with shorter linear or parabolic segments \cite{kallmann2003planning}. It can generate highly-smooth trajectories but does not perform well in the higher-order case.
\cite{rastelli2014dynamic} uses B\'{e}zier Curves to smooth the trajectory depended on control points with low computational time but lacks malleability when the curve degree increases.
\cite{shiller1991dynamic} introduces B-spline to generate a smooth and continuous curvature trajectory, which is a piece-wise polynomial parametric curve. 
\cite{pan2012collision} smooths piece-wise linear collision-free trajectories after sample-based planners computing and uses local spline refinement to avoid collisions. 
The latter one defines object functions about different constrained variables and minimizes or maximize them to produce expected trajectories.
\cite{ziegler2014trajectory} achieves $C_2$ continuous by designing a function that takes the velocity, acceleration, and jerk into account, but not considering the dynamics constraints.
\cite{mellinger2011minimum} proposes the Minimum Snap method, which generates a safe and smooth trajectory by minimizing a cost function derived from the snap and accelerations, satisfying dynamics constraints. It is mainly applied but not limited to aerial vehicles and is also available for other differential systems. 
As the learning methods develop, the method of using machine learning to learn models via data from a manual operation is presented in \cite{lupashin2010simple}, and \cite{abbeel2008apprenticeship} improves the performance by reinforcement learning. However, learning techniques do not deal with high-dynamics environments well with obstacles around. 
Moreover, safety is also critical for trajectory generation, which means collision-free and dynamically feasible. \cite{kim2002nonlinear} proposes an MPC-based method under input and state constraints. However, this method is only efficient when the linearized model is fully controllable or if a control Lyapunov function can be synthesized \cite{shen2017trajectory}. 
In conclusion, it is of vital importance and challenging to generate smooth and collision-free trajectories for a non-linear and under-actuated dynamics system.

%% file: section/03_Methodology.tex
\section{Methodology}

\subsection{ASV Dynamics}

\begin{figure}[htbp]
\centering
\includegraphics[width=0.6\linewidth]{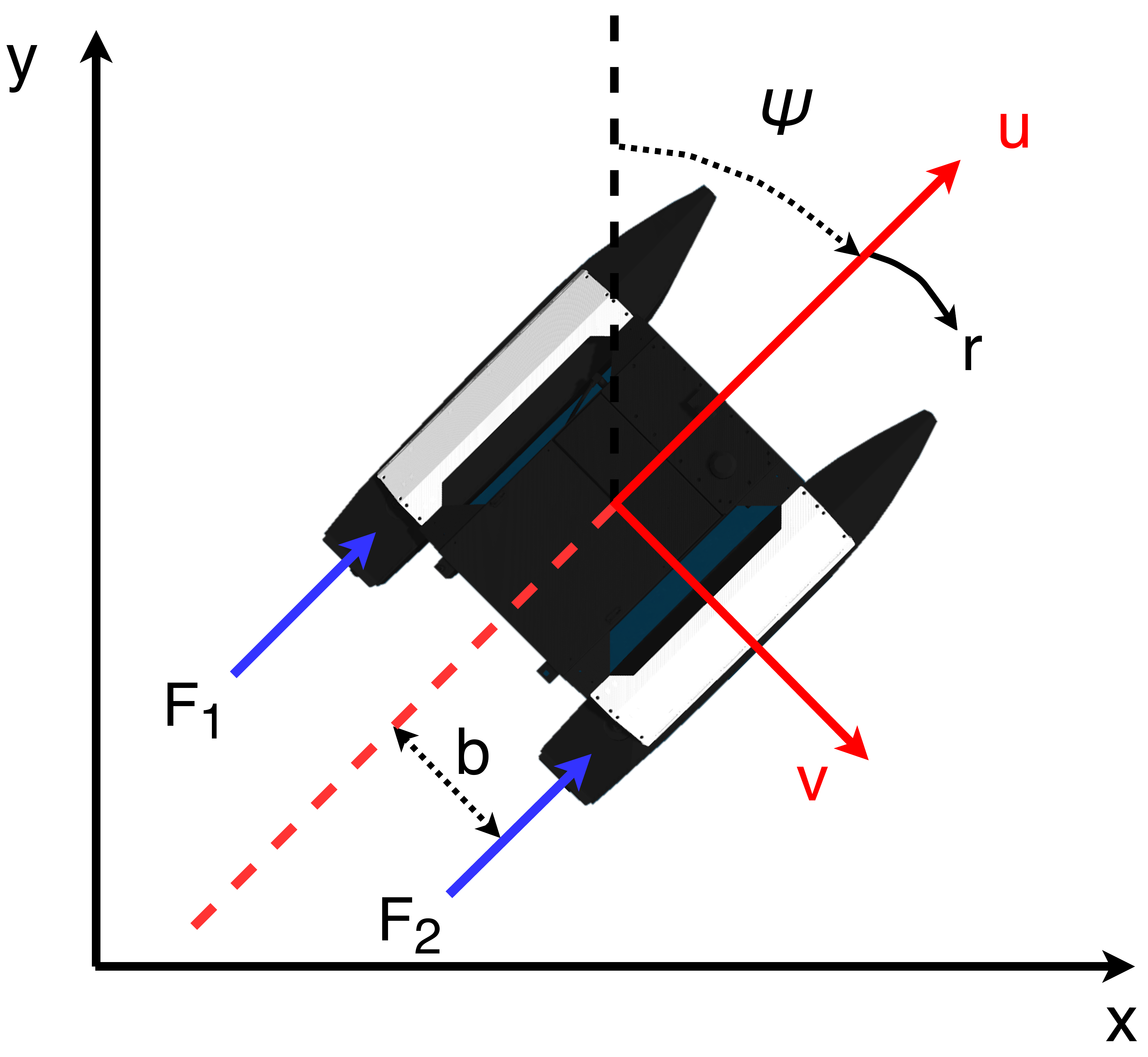}
\caption{Under-actuated ASV module, x-y denotes earth-fixed reference frame and u-v denotes body reference frame}
\label{asv_dynamic}
\end{figure}
In this paper, we use Kingfisher robot from Clearpath Robotics (see also \cite{Clearpath.com}) as ASV model. It's an agile, battery operated boat with two propeller engines designed for research and rapid prototyping.

Two reference frames are introduced when characterizing ASV kinodynamics. The earth-fixed reference frame use North-East coordinates fixed to the ground, while the body reference frame follows the centroid of ASV. From \cite{fossen2011handbook} \cite{mccue2016handbook}, the kinodynamics equations of ASV shows in Fig.\ref{asv_dynamic} are
\begin{equation}\label{dy_eq}
\left\{\begin{array}{ll}
{\dot{\eta}} & {=R(\psi) \nu} \\
{M \dot{\nu}} & {=\tau-C(\nu) \nu-D(\nu) \nu}
\end{array}\right.
\end{equation}

Let $\eta=\left[{x} \quad{y} \quad{\psi} \right]^{T} \in \mathcal{R}^{3}$ and $\nu=\left[{u} \quad{v} \quad {r}\right]^{T} \in \mathcal{R}^{3}$ denote pose vector and velocity vector, respectively. $x$ and $y$ are earth-fixed reference frame coordinates, and $\psi$ represents the anticlockwise angle from earth-fixed frame to body reference frame, also called heading angle. $u$ and $v$ denote velocities in surge and sway, respectively. $r$ is the yaw rate. $R(\psi)$ is the rotation transform matrix from body-fix reference frame to earth-fix reference frame. $M$ is the intertial matrix, $C(\nu)$ denotes a Coriolis and Centripetal matrix and $D(\nu)$ denotes the damping matrix.

\begin{equation*}
\begin{aligned}
R(\psi)=&\left[\begin{array}{ccc}
{\cos (\psi)} & {-\sin (\psi)} & {0} \\
{\sin (\psi)} & {\cos (\psi)} & {0} \\
{0} & {0} & {1}
\end{array}\right] \\
\end{aligned}
\end{equation*}
\begin{equation*}
\begin{aligned}
M =&\left[\begin{array}{ccc}
{m} & {0} & {0} \\
{0} & {m} & {0} \\
{0} & {0} & {I_{z}}
\end{array}\right] \\
\end{aligned}
\end{equation*}
\begin{equation*}
\begin{aligned}
C(\nu)=&\left[\begin{array}{ccc}
{0} & {0} & {m v} \\
{0} & {0} & {-m u} \\
{m v} & {-m u} & {0} \\
\end{array}\right] \\
\end{aligned}
\end{equation*}
\begin{equation*}
\begin{aligned}
D(\nu)=&\left[\begin{array}{ccc}
{X_u} & {0} & {0} \\
{0} & {Y_v} & {0} \\
{0} & {0} & {N_r} \\
\end{array}\right]
\end{aligned}
\end{equation*}

The jerk of ASV can be calculated using model dynamic equation \eqref{dy_eq} as
\begin{equation}
\left\{ 
\begin{aligned}
\ddot{u} &=\frac{\dot{\tau_{u}}}{m}-v \dot{r}-\dot{v} r - \frac{X_{u}}m \dot{u} \\
\ddot{v} &=\dot{u} r + u \dot{r}-\frac{Y_{v}}{m} \dot{v} \\
\ddot{r} &=\frac{\dot{\tau_{r}}}{I_{z}}-\frac{N_{r}}{I_{z} } \dot{r}
\end{aligned}
\right.
\end{equation}

For the under-actuated ASV module discussed in this paper, two motor thrusts are incapable of providing direct control in sway velocity. Thus the control input $\tau = \left[\tau_u \quad \tau_v  \quad \tau_r  \right]^{T} \in \mathcal{R}^{3}$ can be represented by linear combination of two thrust forces, as $b$ denotes half of ASV width:
\begin{equation}
\left\{ 
\begin{array}{l}
{\tau_{u}=F_{1}+F_{2}} \\ 
{\tau_{v}=0} \\
{\tau_{r}=b\left(F_{1}-F_{2}\right)}
\end{array}
\right.
\end{equation}

Unlike other researches usually use position, direction and velocities as state variable  $\mathbf{x}$, we also add acceleration into state variable for trajectory optimization in the following section. So the state variable and control input $\bm{\tau}$ are:

\begin{equation}
\begin{aligned}
    \bm{x} &=\left[{x} \ {y} \ {\psi}\ {u}\ {v} \  {r} \ \dot{u} \  \dot{v} \  \dot{r} \right]^{T} \\
    \bm{\tau}& =  \left[ \tau_u \ \tau_r \right]^{T}
\end{aligned}
\end{equation}

\subsection{Two-Stage Trajectory Planning}
 
\begin{figure*}[!t]
\centering
\includegraphics[width=\linewidth]{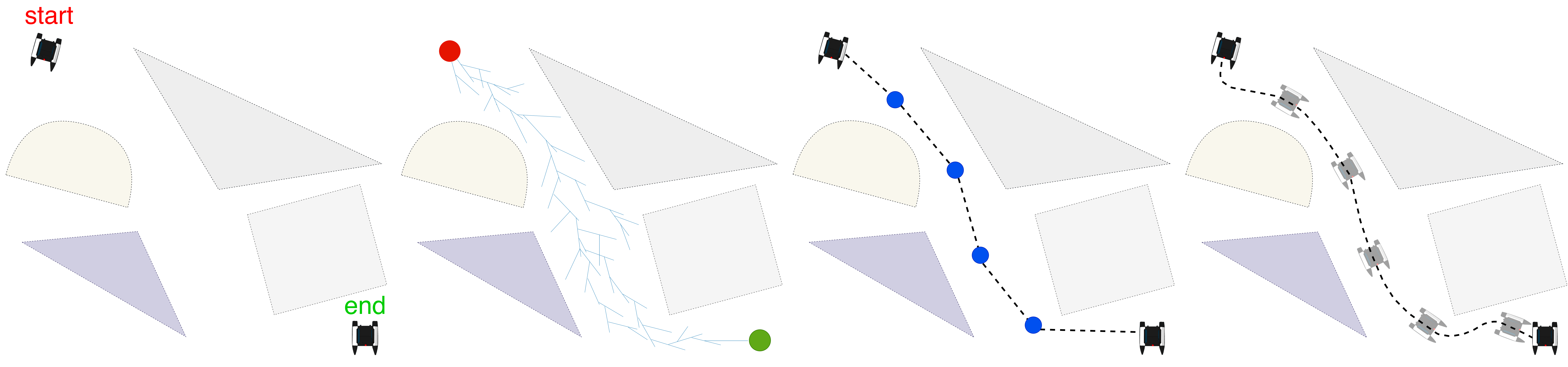}
\caption{(a). The  given map with obstacles and demonstrate location of initial and target point; (b). The front-end RRT* searches the feasible path from the initial point to the target destination ; (c). Several waypoints are extracted from previous path through obstacles; (d). The proposed method generates a collision-free and smooth trajectory}
\label{path}
\end{figure*}

In this paper, we propose a framework, which is divided into two sections: the front-end and back-end. The first step produces several collision-free waypoints in obstacle-occupied environments using path-searching methods. The second step generates a smooth and time-adjusting trajectory accounting for dynamics and state constraints. Considering limited energy while performing tasks, we design the Minimum Control Input Function to save energy. Non-smooth trajectory brings unnatural and even unsafe motions, so we design the Minimum Acceleration Function to sail as smoothly as possible. 

We suppose that those obstacles in the target area are known and fixed before the trajectory planned. As Fig.\ref{path} shows, given the initial position, the target destination, and a map with known obstacles, the front-end produces a collision-free trajectory through the waters with obstacles. This trajectory constrains $N$ waypoints, and each of them has exact coordinates and is away from obstacles. Considering computational-efficiency and optimality, we use RRT* planner \cite{karaman2011sampling}\cite{karaman2010optimal} as front-end path searching module. As Algorithm \ref{alg:A} shows, RRT* iteratively builds the tree by randomly sampling the state $s_{rand}$ from the configuration space and choose a path $x_{new}$, which extends the closest node $s_{nearest}$ in the tree toward the sample. If there is no obstacle lying in this path, RRT* compares the neighbors of $s_{new}$ and evaluates the cost of each node as the parent node instead of directly inserting the new node $s_{new}$ into the tree. The node with the lowest cost becomes the parent node and is added to the tree. The total cost contains the cost of reaching the potential parent node and the cost of the path to $s_{new}$. This process iteratively modifies the tree, reduces the total cost and makes the results asymptotic optimal.

\begin{algorithm}
\caption{Rapidly-exploring Random Tree*}
\label{alg:A}
\begin{algorithmic}[1]
\STATE $T \leftarrow $ InitializeTree()
\STATE $T \leftarrow $ InsertNode($\emptyset, s_{init}, T$)
\FOR{$i=1$ to $N$}
\STATE $s_{rand} \leftarrow$ Sample($i$)
\STATE $s_{nearest} \leftarrow$ Nearest($T, s_{rand}$) 
\STATE $(x_{new},u_{new},T_{new}) \leftarrow$ Steer($s_{nearest}, s_{rand}$)
\IF{CollosionFree($s_{nearest}, s_{rand}$)} 
\STATE $S_{near} \leftarrow$ Near($T, s_{new}, |V|$)
\STATE $s_{min} \leftarrow s_{nearest}$
\STATE $c_{min} \leftarrow$ Cost($s_{nearest}$) + c($s_{new}$)
\FOR{$s_{near} \in S_{near}$} 
\STATE $(x',u',T') \leftarrow$ Steer($s_{near}, s_{new}$)
\IF{CollosionFree($x'$) and $x'(T')=s_{new}$}
\STATE $c' = Cost(s_{near}) + c(x')$ 
\IF{$c' < Cost(s_{near})$ and $c' < c_{min}$}
\STATE $s_{min} \leftarrow s_{near}$
\STATE $c_{min} \leftarrow c'$
\ENDIF
\ENDIF
\ENDFOR 
\STATE $T \leftarrow$ InsertNode($s_{min}, s_{new}, T$)
\STATE $T \leftarrow$ Rewrite($T, s_{near},s_{min},s_{new}$)
\ENDIF
\ENDFOR
\RETURN $T$
\end{algorithmic}
\label{RRT_algo}
\end{algorithm}

After the front-end complete generating waypoints, we need to optimize the trajectories which meet the vehicle's dynamic constraints and stay collision-free. Waypoints generated by sample-based planners sometimes result in jerky or unnatural motions, which may cause over-actuation or side sift. It is difficult for an actuator to execute such a path. In order to generate high-quality trajectories, we optimize each sub-trajectory considering the ASV's dynamics and state's constraints. Suppose $N$ waypoints generated after front-end, and then there are $N+1$ sub-trajectories considering both initial and target points. It is vital to allocate time for each sub-trajectory to achieve time-optimal. In terms of saving energy, we propose an object function accumulating the input control and time cost. In terms of improving smoothness, we propose another objective function accumulating acceleration and time cost. Meanwhile, the state's constraints are taken into account, including dynamics constraints and sailing corridor constraints. Sailing corridor constraint is a novel method we propose in this paper, applied to ensure the trajectory collision-free.  

%% file: section/04_TrajectoryGeneration.tex
\section{Trajectory Optimization}

\subsection{Optimization Problem}
In order to sail from the initial point (usually the current location) to the target destination and avoid all the obstacles, the trajectory must pass all the waypoints provided by front-end path searching. It should be clear that the state of the start/end point is consistent with our requirements, including position, heading angle, velocity, and acceleration on each axis. When passing the waypoints, we do not make too many requirements on heading angle and velocities.   

Suppose we receive $N$ waypoints from the front-end, the whole trajectory is divided into $N+1$ sub-trajectories. Our optimization object makes the trajectory begin with the initial state $\bm{z}_0$, passing through all waypoints and finally reaches the target state $\bm{z}_f$. Furthermore, the whole trajectory should satisfy all kinodynamic constraints.

\begin{figure}[htbp]
\includegraphics[width=0.9\linewidth]{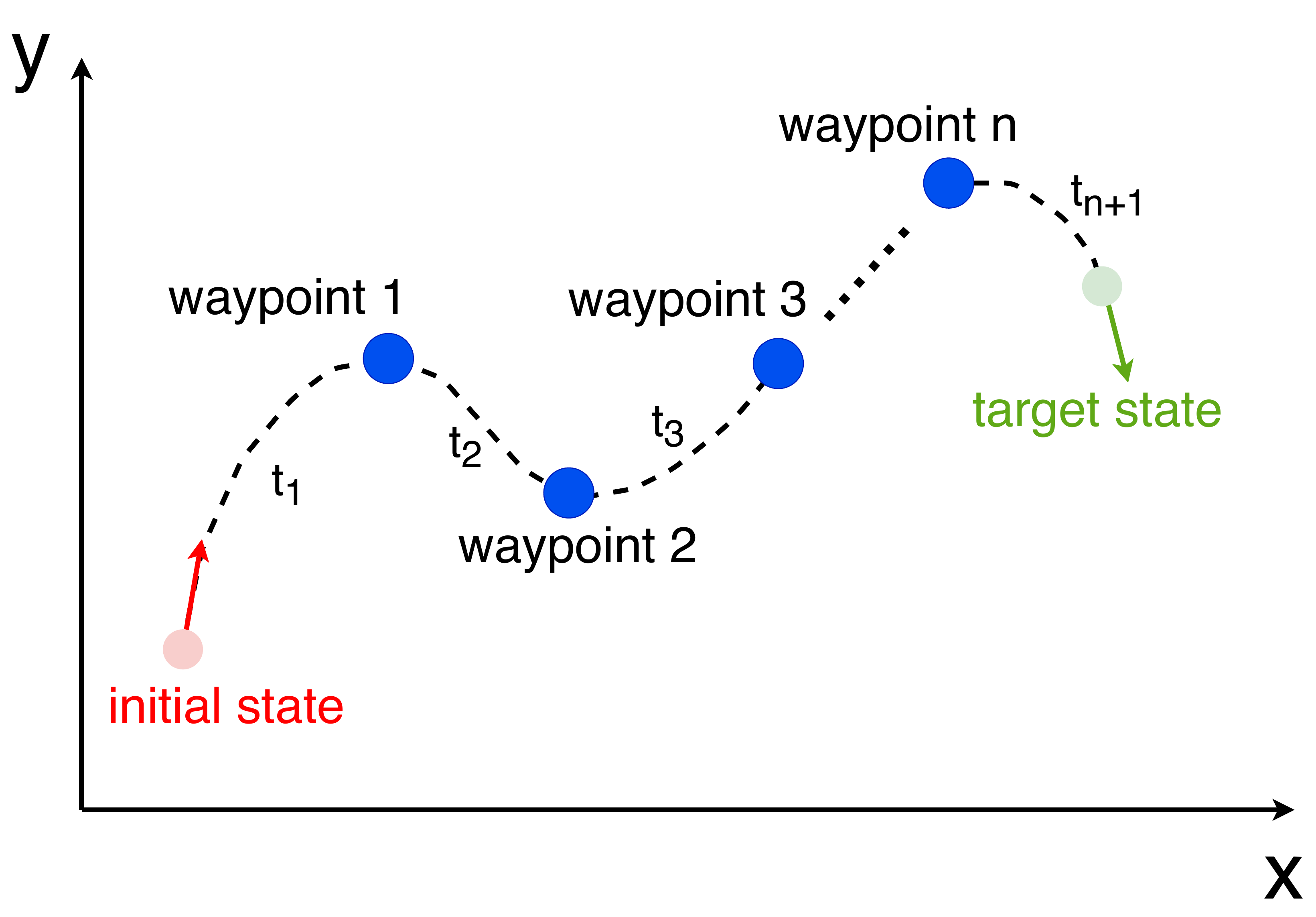}
\caption{Optimize each sub-trajectory and allocate time optimally, blue points denote $n$ waypoints, pink point and green point denote initial and target state, respectively}
\label{waypoint}
\end{figure}

As Fig.\ref{waypoint} shows, for the $i$th ($i \leq N+1$) sub-trajectory, $t_i$ denotes time cost for ASV to complete this sub-trajectory. $\bm{x}_i(t)$ and $\bm{\tau}_i(t)$ denotes the state function and control input function respectively at current sub-trajectory. Obviously, the $i$th sub-trajectory starts from $(i-1)$th waypoint and ends with $i$th waypoint, as the first sub-trajectory starts from the initial point and the last ends with the target point.

Let $J_i$ be the objective function of $i$th sub-trajectory. We propose two different objective functions:

\subsubsection{Minimum Control Input Function}

The most challenging problems for autonomous surface vehicles always contain using limited fuel patrolling as long as possible. This reminds us that it is necessary to plan a fuel-saving trajectory for practical tasks. Two thrusts generate forces $F_1,F_2$ which exert influence on fuel consumption directly and $\bm{\tau}=\left[ \tau_u,\tau_r \right]^T$ are linear combinations of two forces. Minimizing control input $\bm{\tau}$ guarantees ASV sailing in an eco-friendly way along the output trajectory.

Besides, the time factor should be taken into account, as we expect ASV to consume less time while ensuring the task is completed well. Thus optimization problem is as follow:

\begin{equation}
\label{mcif}
\begin{aligned}
    min\  J_i =& \int_{0}^{t_i}\left\|{\bm{\tau}_i}(t)\right\|^2 + \lambda_1 t^2 d t \\
    s.t.\quad& (\ref{timec}) , \ (\ref{velc}) , \   (\ref{ctrlc}) , \  (\ref{corridorc})
\end{aligned}
\end{equation}
where $\lambda_1$ plays a normalization factor.

\subsubsection{Minimum Acceleration Function}
Though minimum control input has advantages in most scenarios, the ASV tends to sailing straightly as well as turning at a small radius. When smoothness becomes a major concern for trajectory planning, minimizing acceleration objective function is a more reasonable choice. Acceleration $\bm{\dot \nu}$ is included in state variables $\bm{x} = [\bm{\eta} ;\bm{\nu};\bm{\dot \nu} ]$. Similar to \eqref{mcif}, time factor is also added into objective function.

\begin{equation}
\begin{aligned}
    J_i =&  \int_{0}^{t_i}\left\|{\bm{\dot{\nu}_i}}(t)\right\|^2 + \lambda_2 t^2 d t \\
     s.t.\quad& (\ref{timec}) , \ (\ref{velc}) , \   (\ref{ctrlc}) , \  (\ref{corridorc})
\end{aligned}    
\end{equation}
where $\lambda_2$ is a normalization factor.

\subsection{Trajectory Constraints}
Apart from dynamic equations and objective functions, some constraints are also required in our problem.

For the $i$th sub-trajectory, 
\subsubsection{Time constraint} 
\begin{equation}
\label{timec}
   t_i \leq  T
\end{equation}
where  $T$ is the maximum time for each sub-trajectory.

\subsubsection{Velocity constraint and acceleration constraint} 
\begin{equation}
\label{velc}
\begin{aligned}
    &\bm{\nu}_{min} \leq \bm{\nu}_i(t) \leq   \bm{\nu}_{max} \\
    &\bm{a}_{min} \leq \bm{\dot \nu}_i(t) \leq   \bm{a}_{max}
\end{aligned}
\end{equation}

\subsubsection{Control input constraint} 
\begin{equation}
\label{ctrlc}
  \bm{\tau}_{min} \leq \bm{\tau}_i(t) \leq   \bm{\tau}_{max} 
\end{equation}

\subsubsection{Sailing corridor constraint}
\begin{figure}[htbp]
\centering
\includegraphics[width=0.75\linewidth,trim=0cm 0cm 0cm -1cm]{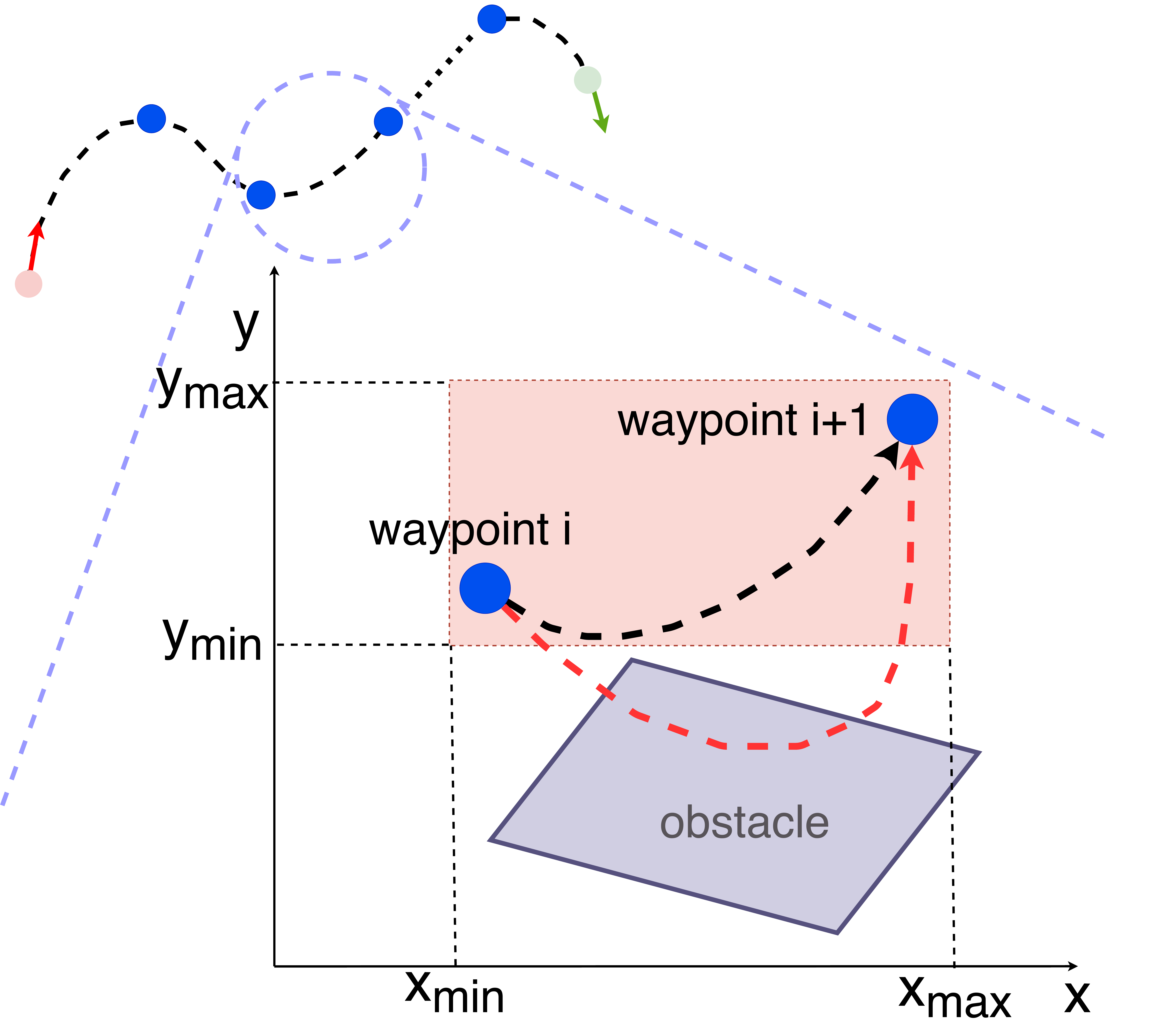}
\caption{The trajectory (black one) with sailing corridor constraints stays collision-free, while the other (red one) collides with the obstacle.}
\label{corridor}
\end{figure}

We also need to consider the constraints resulting from obstacles. Though the front-end algorithm generates waypoints from a collision-free path, the output path connec ts two adjacent waypoints directly, without considering kinodynamics. As the situation in Fig.\ref{corridor} shows, there is no obstacle lies between the $i$th waypoint and the $(i+1)$th waypoint. However, the generated sub-trajectory is subjected to kinodynamics constraints and the initial state at the $i$th waypoint. We suppose ASV has a velocity at a negative y-axis direction when leaving the $i$th waypoint, which needs to adjust a large angle to arrive the $(i+1)$th waypoint. The red dotted line shows a possible but not feasible trajectory because ASV crashes into an obstacle during the voyage.

We propose a method called sailing corridor, which demarcates an area including two endpoints of the current sub-trajectory while excluding all the obstacles surrounding. Using $\bm{p}_i$ denotes the coordinates of the earth-fixed frame on the $i$th sub-trajectory, sailing corridor constraints can be formatted as:

\begin{equation}
\label{corridorc}
  corridor_{min} \leq \bm{p}_i(t) \leq  corridor_{max} 
\end{equation}

\subsubsection{Waypoints constraint}
Finally, we need to optimize all the sub-trajectories in order and combine them into the final trajectory. As described above, besides the initial state and target state, we only constrain the position(x-y coordinate) on each waypoint. It is worth mentioning that for continuity and smoothness of the whole trajectory, both state variables and control input can not rise sudden changes. The characteristic of system states and control functions should remain continuous on each sub-trajectory, the connection of two adjacent trajectories must have consistent states (including velocities and acceleration) and control inputs.

\begin{algorithm}[!hb]
\caption{Optimize Trajectory}
\label{alg:B}
\begin{algorithmic}[1]
\REQUIRE $T$,$\bm{z}_0$,$\bm{z}_f$,waypoints
\ENSURE $t$,$\bm{x}$,$\bm{\tau}$
\FOR{$i=1$ to $(N+1)$}
\IF{$i = 1$} 
\STATE $\bm{x}_i(0) = \bm{z}_0$
\ELSE
\STATE $\bm{x}_i(0) = \bm{x}_{i-1}(t_{i-1})$
\ENDIF
\IF{$i = N+1$} 
\STATE $\bm{x}_i(t_i) = \bm{z}_f$
\ELSE
\STATE $x_i(t_i) = waypoint_{i}.x$
\STATE $y_i(t_i) = waypoint_{i}.y$
\ENDIF
\STATE  Scan surrounding obstacles obtain sailing corridor
\STATE $t_{i,max} \leftarrow T $
\STATE Set constraints $\bm{\nu}_{min}$ $\bm{\nu}_{max}$ $\bm{\dot \nu}_{min}$ $\bm{\dot \nu}_{max}$ $\bm{\tau}_{min}$ $\bm{\tau}_{max}$
\STATE Set objective function $J_i$
\STATE Optimize ($t_i,\bm{x}_i,\bm{\tau}_i$)
\ENDFOR
\end{algorithmic}
\label{optimize_algo}
\end{algorithm}

The optimize procedure presents in Algorithm \ref{optimize_algo}: using initial state and the coordinate of first waypoint, we obtain the analysis formula of time cost, control input function and state variable function regarding the first sub-trajectory; Then we use the last state of the previous sub-trajectory and the next waypoint to optimize the current sub-trajectory. Repeat the above process to generate each sub-trajectory, and we can finally obtain the whole path by connecting them.

%% file: section/05_ExperimentAndResult.tex
\section{Experiment and Results}
\subsection{Simulation Environment}
In the experiment, we adopt Kingfisher robot \cite{Clearpath.com} as the ASV model and use the Gazebo with unmanned surface vehicle plugin provided by \cite{bsb.github} as simulation environment in this section. According to technical specifications of Kingfisher boat, the parameters for ASV model are as $M=diag(29 \quad 29 \quad 2.8),\quad D=diag(20 \quad20 \quad20) $

\begin{figure}[htbp]
\centering
\includegraphics[width=0.6\linewidth,trim=0cm 0cm 0cm -1cm]{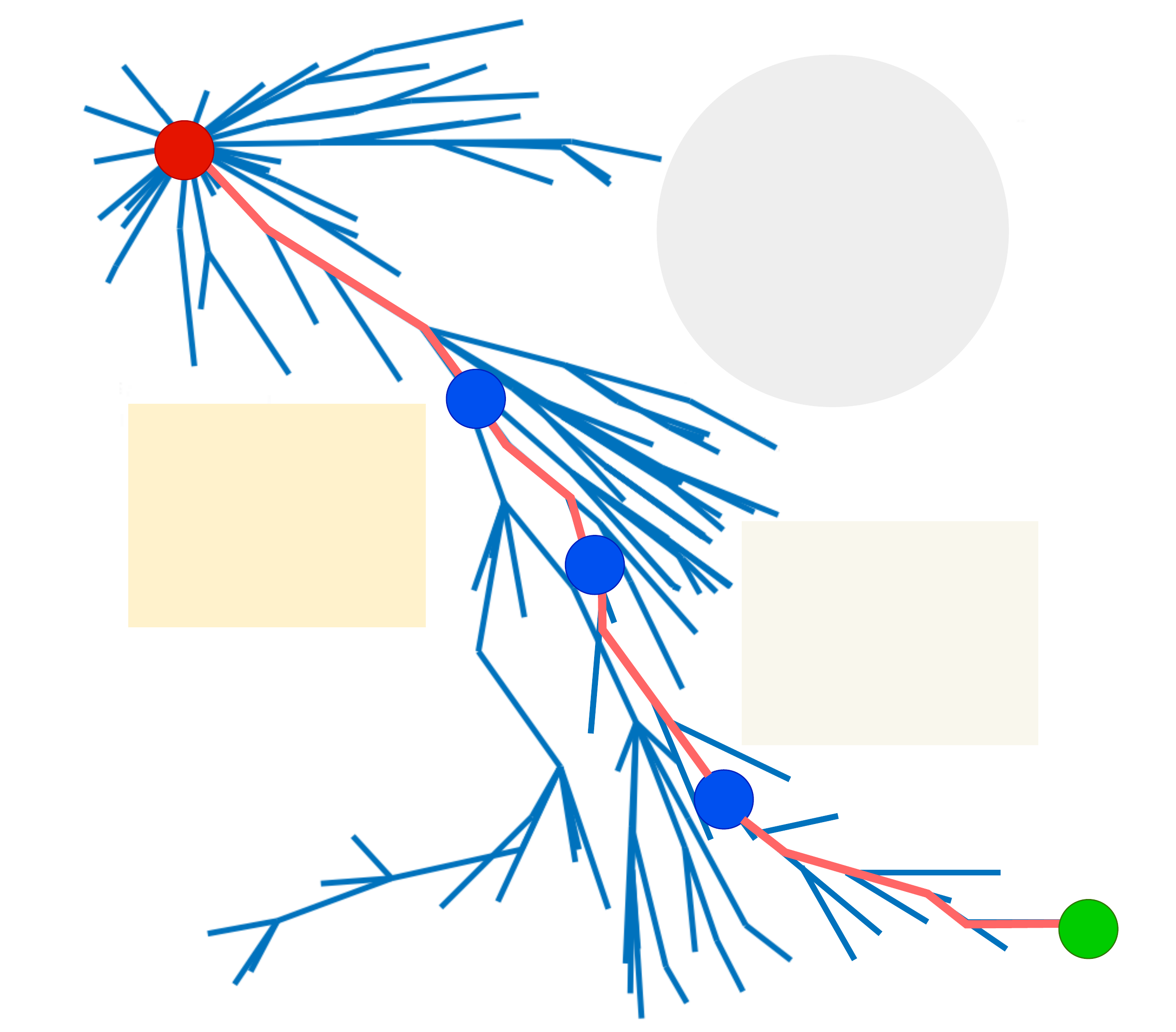}
\caption{RRT* produces a collision-free path from start point to target, and several feasible waypoints are extracted from that path}
\label{rrt}
\end{figure}

As for trajectory optimization method, trapezoidal direct collocation described in \cite{betts2010practical}\cite{kelly2017introduction} is used. We apply a low-order spline to approximate the optimal trajectory. To be specific, the dynamics, objective function and control trajectories are approximated using a linear spline, and the state trajectory is a quadratic spline obtained by integration of the dynamics spline. The integration of a linear spline is computed using the trapezoid rule. 

We place 3 different obstacles (two as cuboid and one as a cylinder) compactly in a $20 \times 20$  square meters task area as shown in Fig.\ref{simu}. Our goal is to make the ASV sailing from the top left corner to the bottom right corner with no collision to any of the obstacles. Both two cuboids have 5 meters in length and 4 meters in width centering at $(4.7,8)$ and $(15,6)$, respectively. We set the start point at $(2,15)$ where ASV heading east with zero velocities and accelerations. We demand ASV still heading east when it arrives at the endpoint $(18,1)$.

\subsection{Simulation Result}

Firstly we provide all the obstacles information and locations of start/goal point to front-end $RRT^*$ planner. After having iterated hundreds of times, the front-end planner has successfully found a feasible path. Three waypoints are extracted from such path which are located at $(8,10)$,$(10,5)$,$(12,3)$, respectively, as shown in Fig.\ref{rrt}.

\subsubsection{Constraints Setting}

As described in Section \uppercase\expandafter{\romannumeral4}, these waypoints along with two endpoints divides the whole path into four sub-trajectories. Now we present the constraints we adapt. Firstly sailing corridor constraints are considered with the information of obstacles. This constraint varies from different sub-trajectories due to different relative positions. For the first sub-trajectory we have $x_{1,max}= 11, y_{1,min} = 10$. The second and third sub-trajectory locates between two cuboids, thus the sailing corridors are both $x_{2,min}= 7.2, x_{2,max} = 12.5$ and $x_{3,min}= 7.2, x_{3,max} = 12.5$. The last sub-trajectory just needs to stay away from the right cuboid, as $y_{4,max}=4$. Sailing corridors are also plotted in Fig.\ref{traj1}. We also regularize $\psi_{min} = - \pi,\psi_{max} = \pi$ for heading angle. Kingfisher model has max surge speed at 1.7 meters per second while moves slowly backwards, thus we set $u_{max}=1.7m/s, u_{min}=-0.1m/s$. Due to the under-actuated feature, ASV cannot control the sway velocity directly, thus we regularize $u_{min}= -1.0m/s, v_{max}=1.0m/s$. Besides, $r_{min}= - \frac{\pi}{6},r_{max}= \frac{\pi}{6}$. We set $T = 35s $ for the max allowed time for ASV to travel in each sub-trajectory. 

We use both two optimize objective functions that we introduce above to generate trajectories.

\subsubsection{ Minimum Control Input Objective}

\begin{figure}[htbp]
\centering
\includegraphics[width=0.8\linewidth]{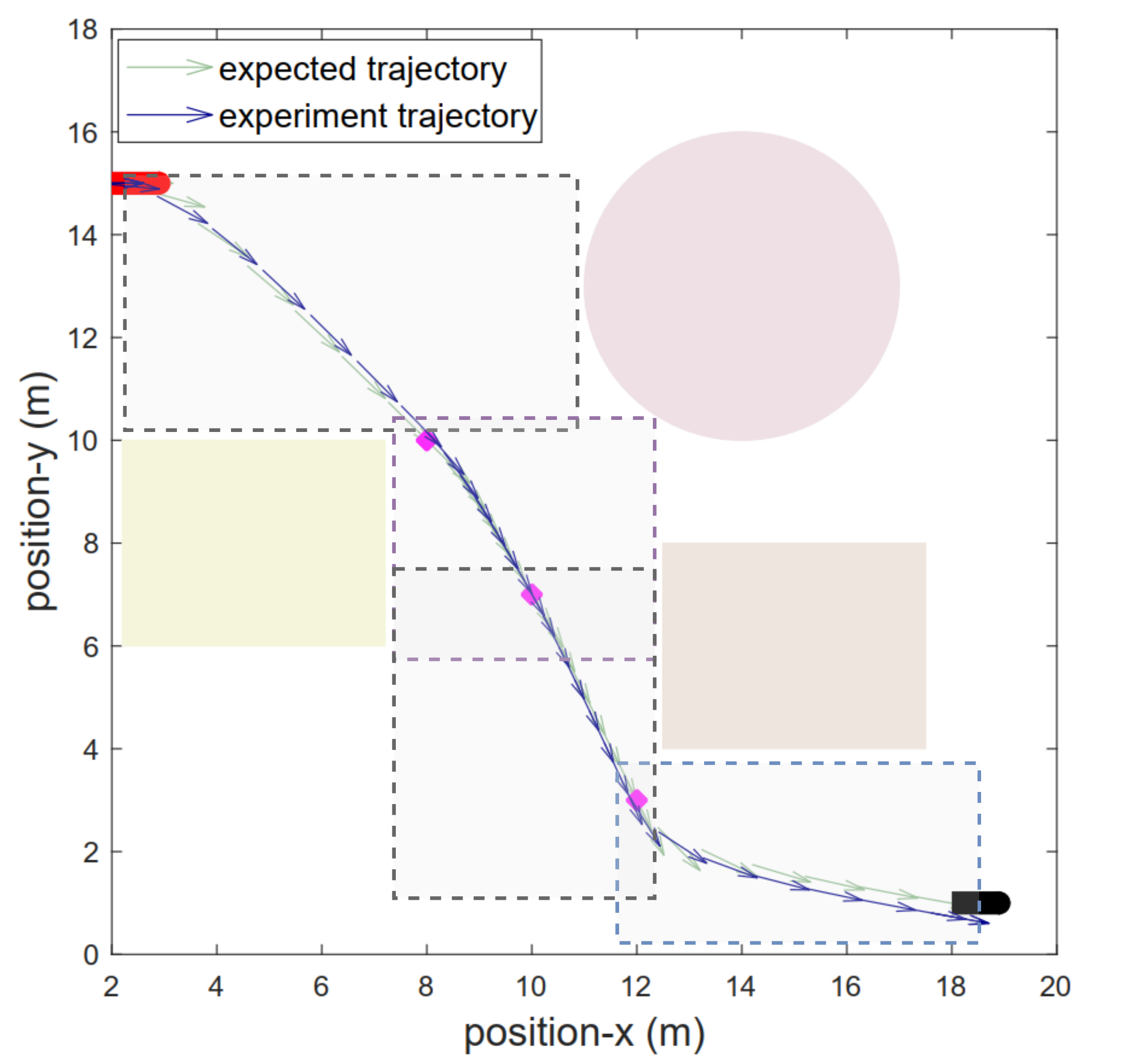}
\caption{Expected and experiment trajectories with the sailing corridor constraints for minimum control input optimization}
\label{traj1}
\end{figure}

\begin{figure}[htbp]
\centering
\includegraphics[width=\linewidth, trim = 1.1cm 1.5cm 0 0]{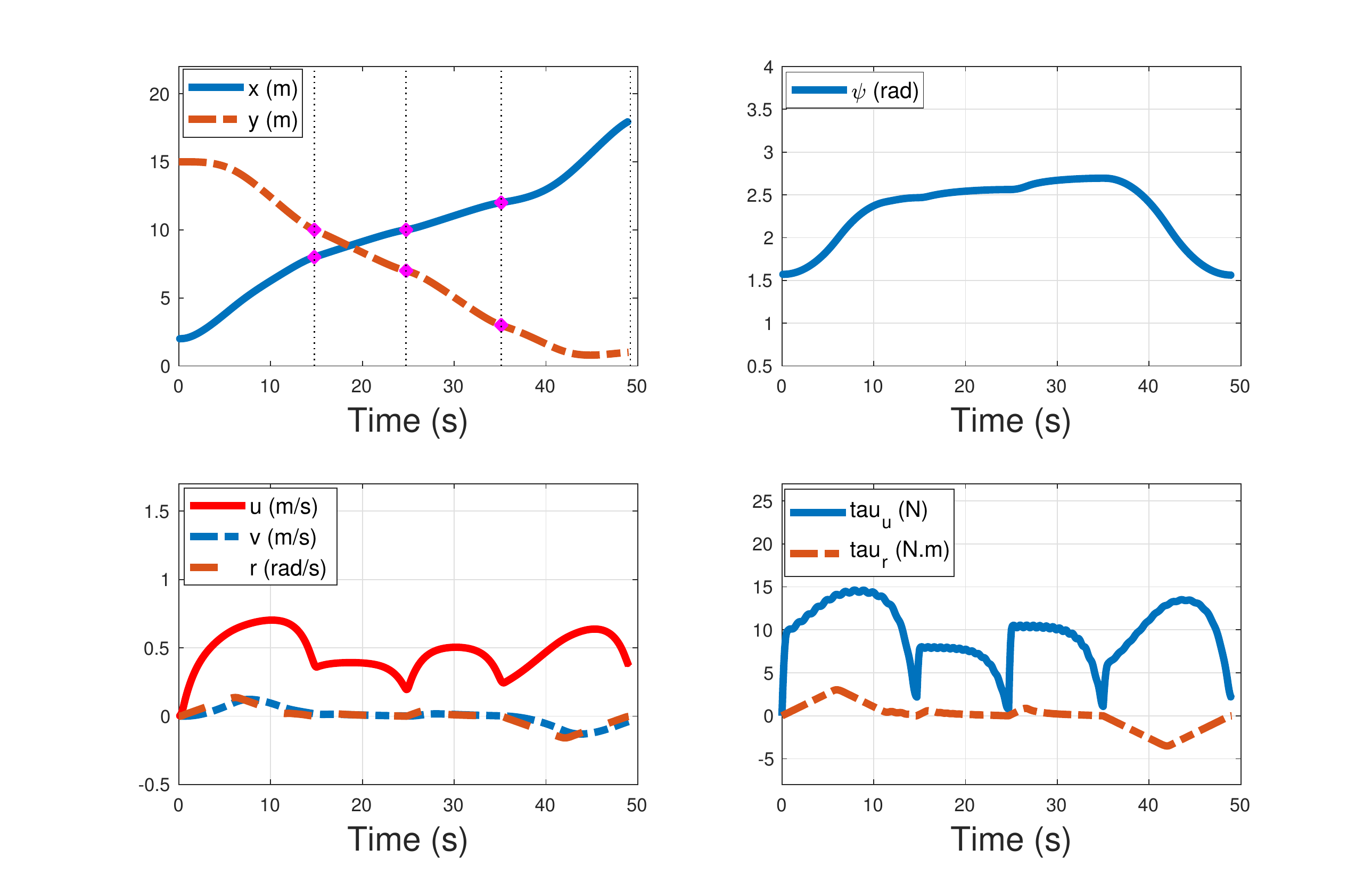}
\caption{Control and state variables for minimum control input optimization; The heading angle $\psi$ are basically consistent with path directions. The velocity on surge $u$ have an average of around $0.5m/s$ along the trajectory and sway velocity $v$ remains low. The control input $\tau_u$ and $\tau_r$ generate cautiously for energy-saving objective. }
\label{example1}
\end{figure}

\begin{figure}[htbp]
\centering
\includegraphics[width=0.8\linewidth,trim=0cm 0cm 0cm -1cm]{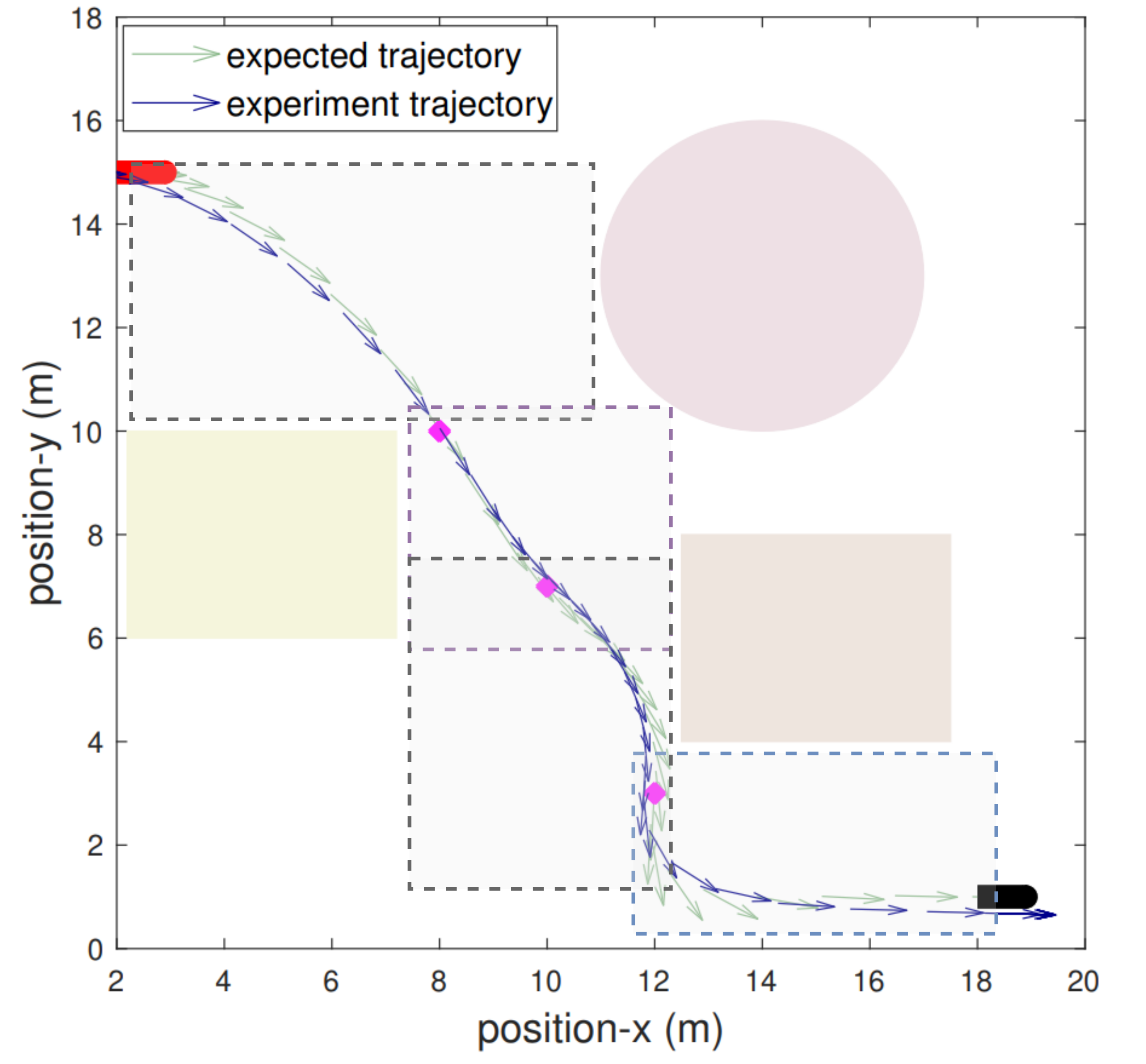}
\caption{Expected and experiment trajectories with the sailing corridor constraints for minimum acceleration optimization}
\label{traj2}
\end{figure}

\begin{figure}[htbp]
\centering
\includegraphics[width=\linewidth, trim = 0.5cm 1.5cm 0 0]{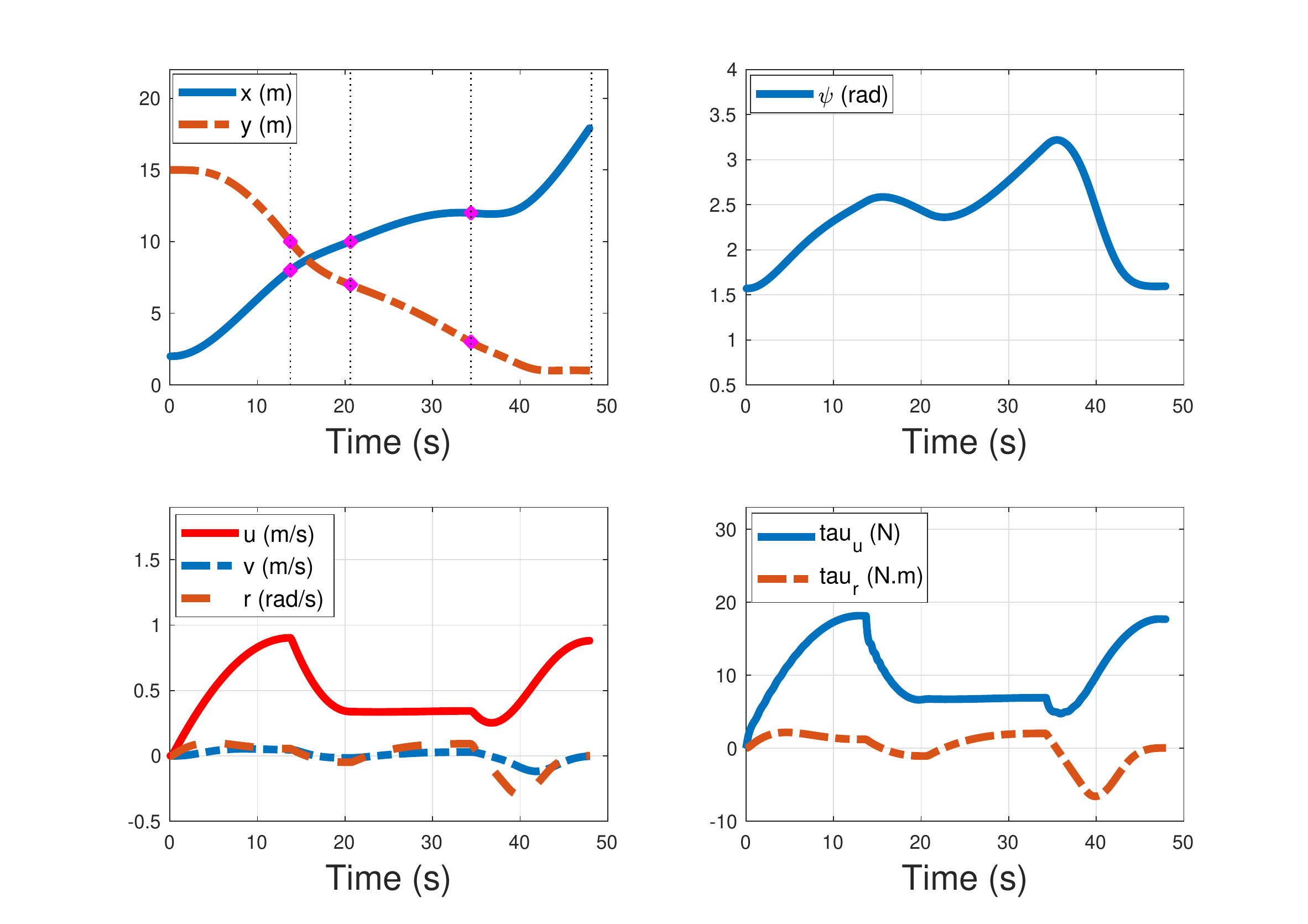}
\caption{Control and state variables for minimum acceleration optimization; The heading angle $\psi$ changes sharply nearby the third waypoint. Velocities $u$, $v$ and $r$ are all inclined to remain constant after reaching relatively high speed. Thus the control input $\tau_u$ and $\tau_r$ are high to keep maintaining such velocities. }
\label{example2}
\end{figure}

The generated trajectory is demonstrated in Fig.\ref{traj1}. The red boat shows the start position and heading angle while the black one presents the end position. Three pink points demonstrate the location of three waypoints generated by the front-end planner. The green arrow sequence denotes the expected trajectory using trapezoidal direct collocation and the blue one denotes the trajectory our model presents in the Gazebo environment. The direction of each arrow shows the heading angle $\psi$ of the ASV at that time.

The detailed control variables and state variables are presented in Fig.\ref{example1}. The generated trajectory spends $t_i = \left[ 14.76 \quad 10.01 \quad 10.36 \quad 14.07 \right]$ seconds on each sub-trajectory and the time duration of $t=49.21$ seconds. We choose the optimization objective as the minimum control input $\bm{\tau}$, ASV tends to have a relatively large $\tau_u$ at the beginning of each sub-trajectory that makes itself have a high surge velocity. Then it needs fewer thrusts to maintain the velocities when getting closer to the end of the sub-trajectory to save fuel.

\subsubsection{ Minimum Acceleration Objective}

The generated trajectory using minimum acceleration objective is demonstrated in Fig.\ref{traj2}. The same symbol represents the same meaning as above.
The detailed control variables and state variables are presented in Fig.\ref{example2}. The generated trajectory spends $t_i = \left[ 13.75 \quad 6.88 \quad 13.70 \quad 13.79 \right]$ seconds on each sub-trajectory and the time duration of $t=48.12$ seconds. Since the optimization objective changes to the acceleration $\bm{\dot \nu}$, the curves of velocities perform smoother than minimum control input in Fig.\ref{example1}. Furthermore, experiment trajectory tends to keep current velocity and disinclines to change directions.

\subsection{Analysis and Comparisons}

Though ASV dynamic model is slightly different between the simulation environment and theoretical model, it still shows that the experiment trajectory can fit the expected one pretty well. As we can see in both Fig.\ref{example1} and Fig.\ref{example2}, the ASV not only arrives at the target with a demanding heading angle but also passes through all the waypoints considering its own width. The waypoints provided by the front-end guide our ship away from obstacles and sailing corridors ensure the safety of the final trajectory.

We compare with the method presented in \cite{wang2018neurodynamics}, and both of the two approaches generate trajectories for under-actuated ASV in the environment with obstacles. The result shown in Wang's article performs relatively low-speed planning with an average speed of $0.269m/s$. Though it manages to avoid two obstacles in the environment, the trajectory shows the ASV sails just at the edge of the giant obstacle which makes it a dangerous and unacceptable trajectory in the real world. As a comparison, the closet distance between ASV and the nearest obstacle in our trajectory generated by minimum input control and minimum acceleration is $1.22m$ and $0.53m$ more than the above one, respectively. It proves the effectiveness and security of our proposed method.

\begin{table}[htbp]
\renewcommand{\arraystretch}{1.3}
\caption{Comparison of Trajectory}
\label{compare}
\centering
\begin{tabular}{|c|c|c|c|}
\hline
 Trajectory &  \makecell[c]{Avg \\Speed(m/s)} & \makecell[c]{Min Obstacle \\ Dist.(m)} & \makecell[c]{Avg Ctrl. \\Input (N/s)} \\
\hline
\makecell[c]{Methods in \cite{wang2018neurodynamics}} & 0.269 & 0 & 3\\
\hline
\makecell[c]{Minimum \\Control Input} & 0.4512 & \textbf{1.22} & 8.3\\
\hline
\makecell[c]{Minimum \\ Acceleration} & \textbf{0.503} & 0.53 & \textbf{10.21}\\
\hline
\end{tabular}
\end{table}

%% file: section/06_Conclusion.tex
\section{Conclusion }
In this paper, we propose a method to generate a collision-free and kinodynamic feasible trajectory for autonomous surface vehicles. We decouple the trajectory planning problem into a front-end feasible path searching and a back-end kinodynamic trajectory optimization. We adopt a sampling-based RRT* path searching to find an obstacle-safe path and extract several waypoints from it. After modeling the type of two-thrusts under-actuated surface vessel, we optimize the position-only path into a trajectory that satisfies the kinodynamic and model constraints. From the perspective of security in the field voyage, we propose a sailing corridor method to guarantee the trajectory away from obstacles.  Moreover, considering limited fuel ASV carrying, we design a numerical objective function which can optimize a fuel-saving trajectory. With trapezoidal direct collocation, all control and state variables are approximated into spline and can be used directly to control the movement of ASV. Finally, we validate our proposed method in a complex simulation environment.